\documentclass[runningheads]{llncs}

\usepackage{url}            
\usepackage{booktabs}       
\usepackage{amsfonts}       
\usepackage{nicefrac}       
\usepackage{microtype}      
\usepackage{xcolor}         
\usepackage{tabularx}
\usepackage{graphicx}
\usepackage{amsmath}
\usepackage{amssymb}
\usepackage{mathtools}
\usepackage{subcaption}
\usepackage{wrapfig}
\usepackage{tabulary,multirow,overpic,xcolor}
\usepackage{amsmath, amssymb, mathtools, bm}
\usepackage{multirow, enumitem, subcaption, float, wrapfig}
\definecolor{mygray}{gray}{0.92}
\definecolor{baselinecolor}{gray}{.9}

\usepackage{listings}
\usepackage{colortbl}
\usepackage{braket}
\usepackage{array}

\usepackage{float}
\usepackage{booktabs}

\newcolumntype{x}[1]{>{\centering\arraybackslash}p{#1pt}}
\newcolumntype{y}[1]{>{\raggedright\arraybackslash}p{#1pt}}
\newcolumntype{z}[1]{>{\raggedleft\arraybackslash}p{#1pt}}

\newcommand{\tablestyle}[2]{\footnotesize \setlength{\tabcolsep}{#1}\renewcommand{\arraystretch}{#2}\centering}

\begin{document}

\title{Rethinking RGB-D Fusion for Semantic Segmentation in Surgical Datasets}

\author{Muhammad Abdullah Jamal and Omid Mohareri}
\institute{Intuitive Surgical Inc.}
\maketitle

\begin{abstract}
Surgical scene understanding is a key technical component for enabling intelligent and context aware systems that can transform various aspects of surgical interventions. In this work, we focus on the semantic segmentation task, propose a simple yet effective multi-modal (RGB and depth) training framework called SurgDepth, and show state-of-the-art (SOTA) results on all publicly available datasets applicable for this task. Unlike previous approaches, which either fine-tune SOTA segmentation models trained on natural images, or encode RGB or RGB-D information using RGB only pre-trained backbones, SurgDepth, which is built on  top of Vision Transformers (ViTs), is designed to encode both RGB and depth information through a simple fusion mechanism. We conduct extensive experiments on benchmark datasets including EndoVis2022, AutoLapro, LapI2I and EndoVis2017 to verify the efficacy of SurgDepth. Specifically, SurgDepth achieves a new SOTA IoU of~\textbf{0.86} on EndoVis 2022 SAR-RARP50 challenge and outperforms the current best method by at least~\textbf{4\%}, using a shallow and compute efficient decoder consisting of ConvNeXt blocks.

\end{abstract}

\section{Introduction}
Intelligent and context aware surgical systems and digital tools have a significant potential to transform minimally invasive procedures by enhancing surgeon and care-team performance, and improving overall safety. Surgical scene parsing is a key component for designing such systems through enabling tasks such as pose estimation~\cite{Pose}, tool tracking~\cite{tooltracking} and phase recognition~\cite{phase}. Applications such as operating room workflow optimization~\cite{workflow}, surgeon skill assessment~\cite{skill1,zia2018surgical} and automation of surgical sub-tasks~\cite{huang2023demonstrationguided} can be built on top of such technologies.

In this paper, we focus on single frame semantic segmentation of instrument, anatomy and other objects present in surgical scenes. The objective is to assign each pixel a correct semantic label. Most of the earlier work~\cite{Shvets_2018,zhao2020learning,MFTAPNet} follow segmentation models built for non-surgical images such as MaskRCNN~\cite{maskrcnn} and UNet~\cite{unet}, either by directly fine-tuning them or incorporating additional cues such as pose~\cite{poseesti}, saliency maps~\cite{saliency}, optical flows~\cite{MFTAPNet} and motion flows~\cite{zhao2020learning}. However, unique challenges such as occlusion, variability in lighting, presence of smoke and blood, and diverse instrument and tissue types limit accuracy, generalizability and clinical translation of present methods. 
Incorporation of 3D geometric information is a promising approach to help enhance the performance of such segmentation algorithms. RGB-D datasets are commonly being used in non-surgical applications such as autonomous driving~\cite{autodrive}, robotics~\cite{robotics} and SLAM~\cite{wang2023coslam}. Existing approaches~\cite{wang2022multimodal,zhang2023delivering} have shown state-of-the-art performance on non-surgical benchmark datasets~\cite{nyuv2,sunrgbd} through methods that leverage both RGB and depth data. However, to the best of our knowledge, very little or no work has been done on effective utilization of RGB-D data for surgical instrument and tissue segmentation. 

\begin{figure}[h]
    \vspace{-10pt}
    \centering
    \includegraphics[width=0.90\textwidth]{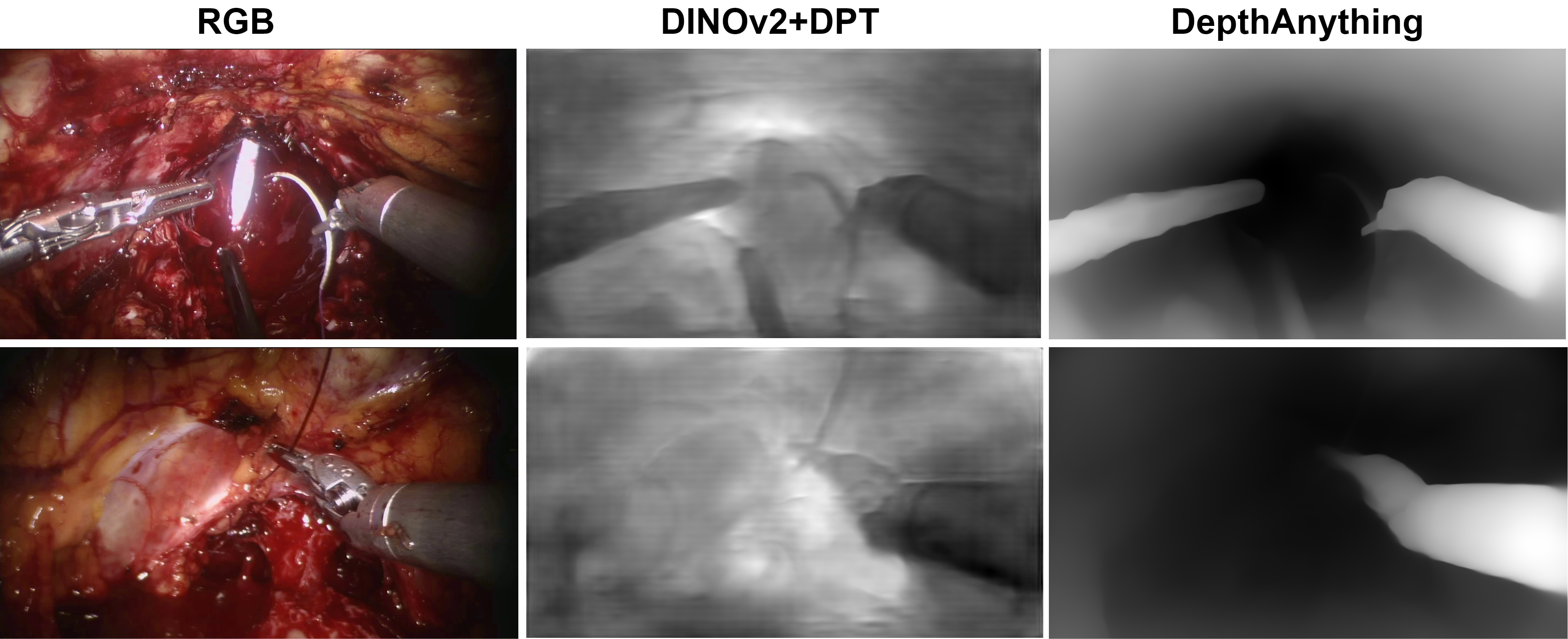}
    \caption{\small Predicted depth maps using DINOv2+DPT~\cite{dinov2} and DepthAnything~\cite{depthanything} on SAR-RARP50 examples.}
    \label{fig:depthmaps}
    \vspace{-10pt}
\end{figure}

This motivates us to present SurgDepth, a simple yet effective RGB-D semantic segmentation framework for endoscopic surgical data. SurgDepth builds the interaction between both data modalities by fusing them using a 3D awareness block. The purpose of this block is to incorporate 3D geometric information from the depth maps to enhance localization of objects and structures. Our fusion module can be plugged in to any Vision Transformer (ViT)~\cite{vit}. Moreover, we propose a shallow decoder based on ConvNeXt~\cite{convnext} blocks to predict the segmentation map. Through extensive experiments, we found out that such method outperforms transformer based decoders such as Segmenter~\cite{strudel2021segmenter}. We employ state-of-the-art depth estimation models like DINOv2~\cite{dinov2} and DepthAnything~\cite{depthanything} to predict depth maps from RGB only surgical videos. Figure~\ref{fig:depthmaps} shows some of the predicted depth maps using these models on SAR-RARP50 dataset~\cite{sarrarp50}.

We demonstrate the efficacy of SurgDepth on multiple datasets such as SAR-RARP50~\cite{sarrarp50}, AutoLapro~\cite{autolaparo}, LapI2I~\cite{lapi2i}, CholecSeg8k~\cite{cholecseg8k} and EndoVis2017~\cite{endovis17}. By adding a ConvNeXt based decoder, we outperform all competing methods while requiring less computational resources. More specifically, our approach achieves \textbf{0.862} IoU on SAR-RARP50 with 98.37M parameters, a new SOTA performance, as compared to Uninades, the best performing method reported in~\cite{sarrarp50}, which achieves 0.829 IoU with 107M parameters. 

\noindent To summarize, our main contributions are:
\begin{enumerate}
    \item A new RGB-D training framework called SurgDepth for semantic segmentation in surgical scenes.
    \item A new 3D awareness fusion block to encode 3D geometric information from depth maps, and a shallow decoder to produce segmentation maps.
    \item Extensive experiments on five benchmark datasets showing new SOTA performance with less computation cost.
\end{enumerate}

\section{Related Work}
There are a few publicly available datasets~\cite{sarrarp50,endovis17,autolaparo} for surgical instrument and tissue segmentation. With the rise of EndoVis challenge, various surgical scene understanding techniques have been explored. In particular, approaches for instrument segmentation can be grouped as semantic~\cite{MFTAPNet,Shvets_2018,PAAD,unetplus} and instance segmentation~\cite{ISINet2020,konginstance,maskrcnn}. Our work targets the semantic segmentation task.

\paragraph{\textbf{Semantic Segmentation.}} TernausNet~\cite{Shvets_2018} used a UNet architecture~\cite{unet} on the top of pre-trained VGG encoder for binary instrument segmentation. \cite{unetplus} proposed a UNet plus architecture, a modified encoder-decoder UNet with data augmentation techniques for medical image segmentation.~\cite{PAAD} proposed a progressive alternating attention network (PAANet) which consists of progressive alternating attention dense (PAAD) blocks to construct attention guided map from all scales. MF-TAPNet~\cite{MFTAPNet} incorporates temporal priors by leveraging motion flow to an attention pyramid network. In addition to the above approaches, ~\cite{realtime1,realtime2} target real-time semantic segmentation.

\paragraph{\textbf{Instance Segmentation.}} Unlike semantic segmentation, which assigns class labels to each pixel, instance segmentation, or mask classification, is an alternative paradigm that assigns class labels to each object instance or binary mask. Most of the earlier work primarily use Mask-RCNN~\cite{maskrcnn}. ISINet~\cite{ISINet2020} builds on the top of Mask-RCNN and proposes a temporal consistency module by taking advantage of the sequential nature of the video data.~\cite{konginstance} re-defined Mask-RCNN by improving region proposal network with anchor optimization. Another line of work in this domain is to develop specialized models~\cite{maskclassify,apmtl}. AP-MTL~\cite{apmtl} proposed an encoder-decoder multi-task learning architecture for real-time instance segmentation.



\section{SurgDepth}
\begin{figure}[t]
    \centering
    \includegraphics[width=0.99\textwidth]{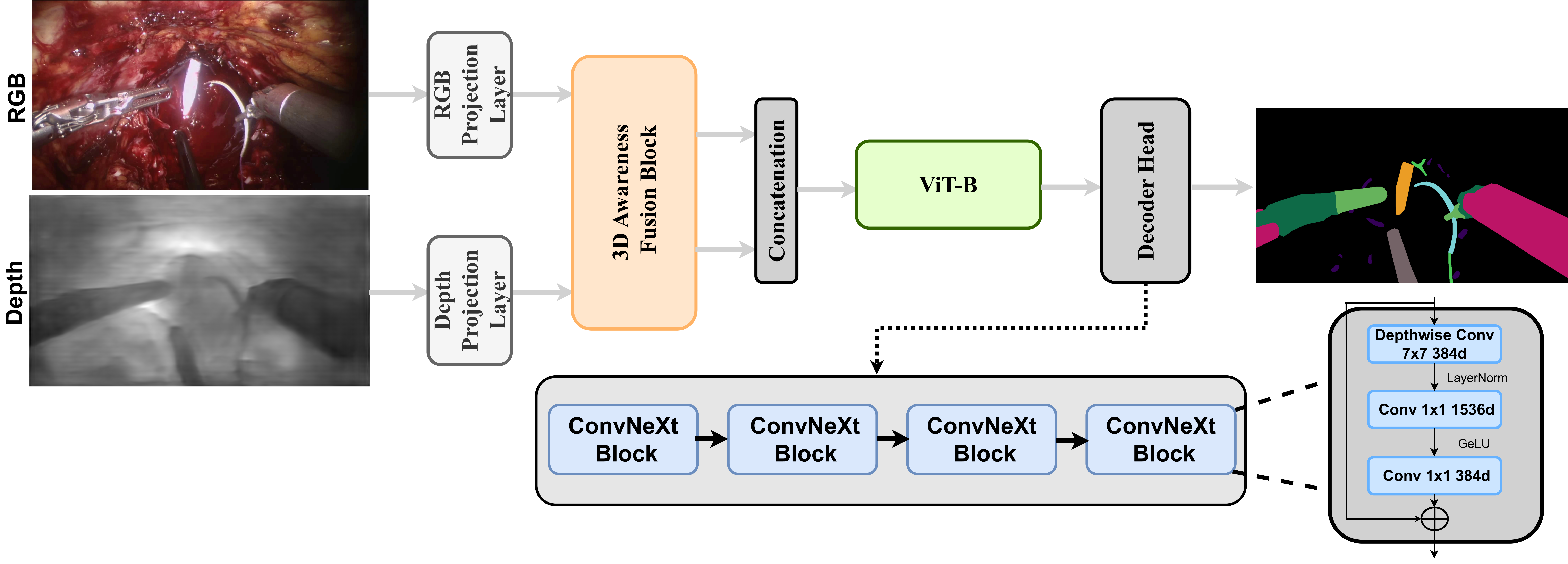}
    \caption{\small Overall architecture of SurgDepth. First, we encode the 3D geometric information using a 3D awareness fusion block and then encode the concatenated RGB-D in ViT-B. Then, the RGB features are passed to a shallow decoder head to predict the segmentation map.}
    \label{fig:approach}
\end{figure}

SurgDepth follows a standard encoder-decoder architecture as illustrated in Figure~\ref{fig:approach}. The goal of the encoder is to learn discriminative representations while the decoder is responsible for transforming these features into segmentation maps.

\noindent An RGB image and depth map with spatial size of \textsc{H $\times$ W} are first processed through modality-specific projection layers consisting of a single convolutional layer. Then, the RGB and depth features are passed to the fusion block which encodes 3D geometrical information. Next, to learn useful representations, the modality-specific features are concatenated and passed to the encoder, which is a ViT in our case. Finally, the features from the encoder are passed to a lightweight decoder to produce the segmentation map of size \textsc{H $\times$ W}. 

\subsection{3D Awareness Fusion Block}
\begin{wrapfigure}{r}{0.33\textwidth}
  \setlength{\abovecaptionskip}{-1pt}
  \setlength{\belowcaptionskip}{-0.3cm}  
  \vspace{-40pt}
  \centering
  \includegraphics[width=0.99\linewidth]{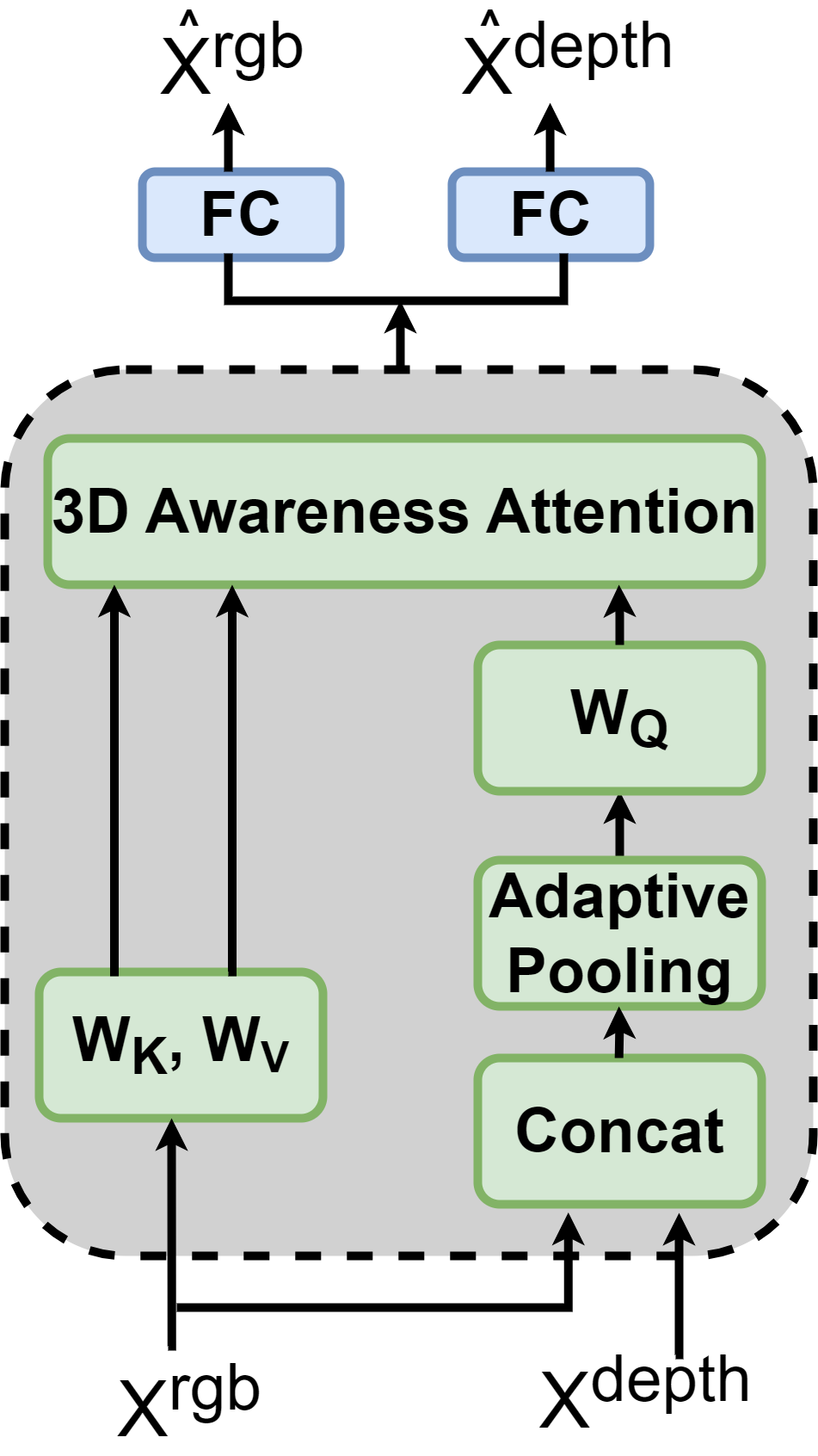}
  \caption{\small 3D awareness fusion block.}
  \vspace{-15pt}
  \label{fig:fusionblock}
\end{wrapfigure}

Our fusion block consists of a 3D awareness attention module that incorporates 3D information from the depth maps to enhance  localization of the semantic classes as shown in Figure~\ref{fig:fusionblock}. Given the RGB $X^{rgb}_{i}$ and the depth $X^{depth}_{i}$ features, we first concatenate the modality-specific features and then down-sample them through an adaptive pooling layer to reduce the computational complexity and generate query (Q) features. The key (K) and value (V) are extracted from the RGB features $X^{rgb}_{i}$. This can be formulated as:

\begin{equation}
\begin{aligned}
   Q = \mathrm{FC}(\mathrm{AdaptivePool}_{k\times k}(Concat(X^{rgb}_{i},X^{depth}_{i}))), \\
   K = \mathrm{FC}(X^{rgb}_{i}), 
   V = \mathrm{FC}(X^{rgb}_{i}),
  \label{eq:Q1}
\end{aligned}
\end{equation}

where $\mathrm{AdaptivePool}$ performs adaptive average pooling to downsample the spatial size to $k\times k$, and $\mathrm{FC}$ is a fully connected layer. Based on the Q, K, and V, we formulate the attention module as:

\begin{equation} 
   X_{fusion} = \mathrm{Bilinear}(V\cdot \mathrm{Softmax} (\frac{{Q}^{\top}{K}}{\sqrt{C^{d}}})),
  \label{eq:GAA1}
\end{equation}

where $\mathrm{Bilinear}(\cdot)$ is a bilinear upsampling operation ($\mathbf{F.interpolate()}$ in Pytorch) that converts the spatial size from $k\times k$ to $h\times w$ and $C^{d}$ represents the dimension of Q, K and V. Finally, the features $X_{fusion}$ are passed to two projection layers (FC) to produce updated RGB features $\hat{X}^{rgb}_i$ and depth features $\hat{X}^{depth}_i$.

\subsection{Overall Architecture}

RGB features $\hat{X}^{rgb}_i$ and depth features $\hat{X}^{depth}_i$ are concatenated and passed through the ViT encoder to encode RGB-D data. The features from the encoder's last layer are passed to the lightweight decoder to yield segmentation maps. We empirically found that passing only RGB features to the decoder yields higher performance as compared to passing both RGB and depth features (c.f. Table~\ref{tab:decoderinput}). Our lightweight decoder consists of ConvNeXt blocks~\cite{convnext} and a convolutional layer for producing segmentation maps. Each ConvNext block has one depth-wise convolutional layer with a kernal size of \textsc{7 $\times$ 7} and two point wise convolutional layers. It follows the inverted bottleneck design where the dimension of the middle point-wise convolution is four times bigger than the input dimension. Please follow~\cite{convnext} for more details on the block. We also experimented with Segmenter~\cite{strudel2021segmenter} as the decoder but empirically found that it doesn't outperform our ConvNeXt based decoder and brings about computational overhead. 

\vspace{-5pt}
\section{Experiments}
We evaluate the performance of our architecture on multiple benchmark datasets.
\subsection{Datasets and Evaluation.}

\paragraph{\textbf{SAR-RARP50 Challenge~\cite{sarrarp50}.}} This dataset contains 50 videos collected from Robot-Assisted Radical Prostatectomy procedures. It consists of 12998 training frames from 40 different videos and 3252 test frames from 10 other videos and nine semantic classes. Please follow~\cite{sarrarp50} for more details on the dataset.
\paragraph{\textbf{AutoLapro~\cite{autolaparo}.}} This dataset has 21 laparoscopic hysterectomy videos recorded with 25 fps and a resolution of 1920×1080 pixels. For segmentation task, the dataset has 1800 frames annotated with nine semantic classes which constitutes 5936 annotations. We use the official  train/validation/test splits provided in the dataset.

\paragraph{\textbf{LapI2I~\cite{lapi2i}.}} The Laparoscopic Image-to-Image (I2I) Translation dataset contains 20,000 synthetic frames annotated with seven classes including liver, fat, diaphragm, tool shaft, tool tip and gallbladder. Synthetic frames are generated using 3D laparoscopic simulations created from CT scans of ten patients. The dataset doesn't provide any official split for train/val/test. We use data from 7 patients in the training set and the rest for testing.

\paragraph{\textbf{CholecSeg8k~\cite{cholecseg8k}.}} This dataset consists of 8,080 laparoscopic cholecystectomy frames  with a resolution of 854 × 480 from 17 video clips in Choloec80. Each image is annotated with 13 classes (abdominal wall,liver, gastrointestinal tract, fat, grasper etc.) at the pixel-level. We use 13 videos in the training set and set aside 4 videos for testing.

\paragraph{\textbf{EndoVis 2017~\cite{endovis17}.}} EndoVis 2017 challenge dataset consists of 10 video sequences of abdominal porcine procedures recorded by a daVinci Xi system. Each video contains 300 frames with a resolution of 1280×1024. The frames are annotated with six instrument classes and an ultrasound probe. For fair comparison, we follow~\cite{Shvets_2018} and use 4-fold cross-validation from 1800 frames (8 x 225). Each fold consists of 1350 and 450 frames for training and validation respectively.

\paragraph{\textbf{Evaluation.}} We use mean intersection over union (IoU) as our evaluation metric ~\cite{strudel2021segmenter,mask2former}.

\subsection{Implementation Details} 
SurgDepth is compatible with any Vision Transformer (ViT) as an encoder in our framework. In this work, we use ViT-B as an encoder, and four ConvNeXt blocks and one convolutional layer to produce segmentation maps. First, we increase the dimensionality \textit{D} of the encoder's output with a linear layer and then reshape it to the size of \textit{H/4} x \textit{W/4} x \textit{D/8}. Before calculating the loss on the segmentation map, we upsample the resolution of the map using bilinear interpolation. We initialize our ViT-B encoder with ImageNet~\cite{imagenet} unsupervised pre-trained weights~\cite{mae}. During training, we use the input resolution of 480x640 across all the datasets. We train the model for 50 epochs with a learning rate of 1e-4 and adamw optimizer. Moreover, we apply random gaussian blurring, random horizontal flip and colorJitter which randomly change the brightness, contrast, saturation and hue of a training image. Lastly, we train our models using the batch size of 2 on 4 NVIDIA A100 GPUs.


\subsection{Results on SAR-RARP50 challenge}
\vspace{-20pt}
\begin{table}[h!]
\centering
\caption {Comparison of SurgDepth with the other approaches on \textbf{SAR-RARP50 Semantic Segmentation}.}
\label{tab:sar-rar} 
\resizebox{0.99\textwidth}{!}{%
\begin{tabular}{c|c|c|c|c|c}
\hline

\hline

\hline
Methods  & Backbone  & Segmenter Head &  Input size & Pre-train  & IoU \\
\hline

\hline

\hline
Hi-Lab 2022    & Swin-B    & SegFormer Ensemble  & 512x512   & N/A & 0.817 \\ \hline

Summer Lab - AI  & Swin-L  & UperNet  & 422x750 & ImageNet-22K  & 0.816  \\ \hline

AIA - Noobs    & EfficientNetB4  &  UNet ++ & 480x640  & ImageNet-1K  & 0.789  \\ \hline

Uninades  & Swin-B   & Mask2Former  & 750x1333 & COCO + EndoVis 17 + EndoVis 18 & 0.829 \\ \hline

MAE~\cite{mae}  & ViT-B   & ConvNeXt  & 480x640 & ImageNet-1K & 0.835 \\ \hline


MAE~\cite{mae}  & ViT-B   & ConvNeXt  & 480x640 & ImageNet-1K + SAR-RARP50 & 0.809 \\ \hline


\textbf{SurgDepth} w/ DINOv2  & ViT-B   & ConvNeXt  & 480x640 & ImageNet-1K & \textbf{0.862} \\ \hline

\textbf{SurgDepth} w/ DINOv2  & ViT-B   & Segmenter  & 480x640 & ImageNet-1K & 0.854 \\ \hline

\textbf{SurgDepth} w/ DepthAnything  & ViT-B   & ConvNeXt  & 480x640 & ImageNet-1K & \textbf{0.858} \\ \hline

\textbf{SurgDepth} w/ DepthAnything  & ViT-B   & Segmenter  & 480x640 & ImageNet-1K & 0.851 \\
\hline

\hline

\hline

\end{tabular}
}%
\vspace{-10pt}
\end{table}

In Table~\ref{tab:sar-rar}, we compare SurgDepth for SAR-RARP50 challenge with the methods that are reported in the~\cite{sarrarp50}. SurgDepth is able to surpass all competitors using the ConvNeXt based decoder. We also compare with the approaches proposed for natural images such as MAE~\cite{mae} which pre-trains ViTs using ImageNet-1k dataset under a masked autoencoding paradigm. As it can be seen in the table, SurgDepth achieves 0.862 IoU, a new state-of-the-art performance on SAR-RARP50 dataset. Compared to best performing method (Uninades) reported in the paper~\cite{sarrarp50}, our approach is computationally more efficient as it requires lesser image resolution and ConvNeXt blocks instead of Mask2Former as segmentation head. Unindaes model has 107M parameters while SurgDepth with ConvNeXt decoder has only 98.37M parameters. 

\subsection{Results on AutoLapro}
\vspace{-20pt}
\begin{table}[h]
\vspace{-10pt}
\centering
\caption {Quantitative comparison on \textbf{AutoLapro Semantic Segmentation}. }
\label{tab:autolapro} 
\resizebox{0.90\textwidth}{!}{%
\begin{tabular}{c|c|c|c|c}
\hline

\hline

\hline
Methods  & Architecture &  Input size & Pre-train  & IoU \\
\hline

\hline

\hline
MaskRCNN~\cite{maskrcnn}    & ResNet-50  & 480x640  & COCO & 67.8 \\ \hline

YOLACT  & ResNet-50 & 480x640 & ImageNet-1K  & 65.2  \\ \hline

YolaactEdge    & ResNet-50 & 480x640  & ImageNet-1K  & 64.4  \\ \hline

MAE~\cite{mae}  & ViT-B + ConvNeXt  & 480x640 & ImageNet-1K & 76.8 \\ \hline


\textbf{SurgDepth} w/ DINOv2  & ViT-B + ConvNeXt  & 480x640 & ImageNet-1K & \textbf{78.0} \\ \hline

\textbf{SurgDepth} w/ DINOv2  & ViT-B + Segmenter  & 480x640 & ImageNet-1K & 76.3 \\ \hline

\textbf{SurgDepth} w/ DepthAnything  & ViT-B + ConvNeXt  & 480x640 & ImageNet-1K & \textbf{77.2} \\ \hline

\textbf{SurgDepth} w/ DepthAnything  & ViT-B + Segmenter  & 480x640 & ImageNet-1K & 76.1 \\
\hline

\hline

\hline

\end{tabular}
}%
\vspace{-10pt}
\end{table}

We compare SurgDepth for AutoLapro dataset with the method reported in ~\cite{autolaparo}. To the best of our knowledge, there hasn't been a lot of work reporting performance on AutoLapro so we mainly reported the results on the methods proposed for natural images. We can see from the Table~\ref{tab:autolapro} that SurgDepth with ConvNeXt decoder outperforms all the baselines. More specifically, it achieves 78.0 IoU setting a new SOTA on AutoLapro dataset. We also demonstrate SOTA results on LapI2I and CholecSeg8k, achieving 98.1 and 55.6 IoU respectively. Please follow supplementary materials for more results on these datasets.

\subsection{Results on EndoVis 2017}
\vspace{-25pt}
\begin{table}[h]
\centering
\caption {Comparison of SurgDepth with the other approaches on \textbf{EndoVis 2017}. }
\label{tab:endovis17} 
\resizebox{0.90\textwidth}{!}{%
\begin{tabular}{c|c|c|c|c}
\hline

\hline

\hline
Methods  & Architecture &  Pre-train  & Challenge IoU & IoU  \\
\hline

\hline

\hline
MaskRCNN~\cite{maskrcnn}    & ResNet-50  & COCO & 45.65  & 41.77 \\ \hline

Mask2Former~\cite{mask2former}  & ResNet-50  & COCO & 40.39  & 39.84\\ \hline

MAE~\cite{mae}  & ViT-B + ConvNeXt  & ImageNet-1K & 55.28  & 44.87 \\ \hline


CascadeRCNN~\cite{cascadercnn}  & ResNet-50  & COCO  & 49.03 & 39.9  \\ \hline

UNetPlus~\cite{unetplus}  & UNet  & None & 36.14  & 13.14 \\ \hline

PlainNet~\cite{MFTAPNet}  & UNet  & None & 36.45  & 13.28 \\ \hline

TernausNet-11~\cite{Shvets_2018} & UNet11  & None & 35.27  & 12.67 \\ \hline

MF-TAPNET~\cite{MFTAPNet} & UNet   & None &  37.35 & 13.49 \\ \hline

ISINet~\cite{ISINet2020}* & ResNet-50   & None &  53.55 & 49.57 \\ \hline

\textbf{SurgDepth} w/ DINOv2  & ViT-B + ConvNeXt  & ImageNet-1K & \textbf{61.93}   &  \textbf{50.29} \\ \hline

\textbf{SurgDepth} w/ DINOv2 (224 $\times$ 224)  & ViT-B + ConvNeXt  & ImageNet-1K & 57.67   & 48.51 \\ \hline

\hline

\hline

\hline

\end{tabular}
}%
\vspace{-10pt}
\end{table}

Table~\ref{tab:endovis17} shows the comparison of SurgDepth with the competing methods on EndoVis 2017 dataset. We report both the challenge IoU and standard IoU. As our approach is a single frame based method, we mainly list the performance of baselines that require single frame in the input. We can see from the table that SurgDepth consistently outperforms all the approaches across both metrics. More specifically, it improves over ISINet by \textbf{15\%} challenge IoU and \textbf{1.45\%} IoU showing the importance of encoding 3D geometric information for semantic segmentation. We want to emphasize that ISINet is an instance-based method for instrument segmentation and we only report the result without the temporal consistency module which takes advantage of the sequential nature of the video data and the instrument motion. With temporal consistency module, ISINet achieves 55.62 challenge IoU which is still lower than SurgDepth but it achieves a higher IoU (52.2), showing that for this dataset, apart from 3D geometric information, other factors like motion, optical flow and temporal information of the instruments are also important.

\vspace{-17pt}






\begin{table}[h]
  \tablestyle{4pt}{1}
  \begin{minipage}{0.49\linewidth}
  \small
  \caption{\small \textbf{Decoder Depth}. SurgDepth performs the best with 4 ConvNeXt blocks in the decoder.}
  \label{tab:decoderdepth}
  \tablestyle{10pt}{1}
  \renewcommand{\arraystretch}{0.8}
  \begin{tabular}{c|c}
    \toprule
    \textbf{Blocks} & \textbf{IoU} \\
    \midrule\midrule
    1 & 0.843 \\
    2 & 0.851 \\
    4 & \textbf{0.862} \\
    8 & 0.856 \\
    \bottomrule
  \end{tabular}\label{tab:rev}
  \end{minipage}
  \small
  \hfill
  \begin{minipage}{0.47\linewidth}
  \vspace{-20pt}
  \small
  \caption{\small \textbf{Decoder input}. SurgDepth performs the best when only RGB features are passed to the decoder. }
  \label{tab:decoderinput}
  \renewcommand{\arraystretch}{0.8}
  \begin{tabular}{lccc} \toprule
    Deocder Input & \textbf{\#Params}  & \textbf{IoU} \\
    \midrule\midrule
    RGB & 98.37M & \textbf{0.862} \\
    RGB+Depth & 103.1M & 0.823 \\
    \bottomrule
  \end{tabular}
  \vspace{-20pt}
  \end{minipage}
\vspace{-20pt}
\end{table}

\subsection{Ablation Study}
\paragraph{\textbf{Input features to the decoder.}} In Table~\ref{tab:decoderinput}, we show that using only RGB features as input to the decoder brings the best performance on SAR-RARP50 dataset as compared to both RGB and depth features while saving the computational cost.

\paragraph{\textbf{Number of ConvNext blocks in decoder.}} Table~\ref{tab:decoderdepth} shows the performance of SurgDepth on SAR-RARP50 challenge by varying the number of ConvNeXt blocks in the decoder. We observe the performance boost when we increase the number of blocks from 1 to 4. However, we see a degradation in the performance when we use a much deeper decoder.
\vspace{-2pt}
\section{Conclusion}
We propose a novel RGB-D training framework called SurgDepth for semantic segmentation in surgical videos. SurgDepth consists of a novel 3D awareness attention block which builds interaction between RGB and depth by incorporating 3D geometric information from the depth maps. The method can be used with any type of Vision Transformer (ViT). Moreover, it can act as a building block for architectures that take video or stereo data as inputs. Our experiments demonstrate that SurgDepth achieves new state-of-the-art performance on five benchmark datasets with less computational cost, thanks to a shallow decoder consisting of ConvNeXt blocks.

\bibliographystyle{splncs04}
\bibliography{references}

\newpage
\appendix
\section*{Appendices}
\addcontentsline{toc}{section}{Appendices}
\renewcommand{\thesubsection}{\Alph{subsection}}

\begin{table}[h]
\centering
\caption {Results on \textbf{LapI2I} and \textbf{CholecSeg8K} Semantic Segmentation.} 
\label{tab:cholec} 
\resizebox{1.0\textwidth}{!}{%
\begin{tabular}{c|c|c|c|c|c}
\hline

\hline

\hline
Methods  & Architecture &  Input size & Pre-train  & LapI2I IoU & CholecSeg8K IoU \\
\hline

\hline

\hline
MaskRCNN~\cite{maskrcnn}    & ResNet-50  & 480x640  & COCO & 90.1 & 47.3 \\ \hline

Mask2Former~\cite{mask2former}  & ResNet-50  & 480x640  & COCO & 93.5 & 48.9 \\ \hline

MAE~\cite{mae}  & ViT-B + ConvNeXt  & 480x640 & ImageNet-1K & 97.4 & 54.1 \\ \hline


\textbf{SurgDepth} w/ DINOv2  & ViT-B + ConvNeXt  & 480x640 & ImageNet-1K & \textbf{98.1} & \textbf{55.6} \\ \hline

\textbf{SurgDepth} w/ DINOv2  & ViT-B + Segmenter  & 480x640 & ImageNet-1K & 96.8 & 54.4 \\ \hline

\hline

\hline

\hline

\end{tabular}
}%
\end{table}

\end{document}